\pdfoutput=1
\documentclass[runningheads]{llncs}
\usepackage{graphicx}

\usepackage{tikz}
\usepackage{comment}
\usepackage{amsmath,amssymb} 
\usepackage{color}

\usepackage[accsupp]{axessibility} 

\usepackage[caption=false]{subfig}

\usepackage{graphicx}
\usepackage[colorlinks=true,pdfborder={0 0 0},linkcolor=blue,citecolor=blue]{hyperref} 
\usepackage{orcidlink} 

\begin{document}
\pagestyle{headings}
\mainmatter
\def\ECCVSubNumber{6667} 

\title{Perspective Phase Angle Model for\\ Polarimetric 3D Reconstruction} 

\titlerunning{PPA Model for Polarimetric 3D Reconstruction}
%

\author{Guangcheng Chen\inst{1,2}\orcidlink{0000-0001-8191-3327} \and
Li He\inst{2,3}\orcidlink{0000-0003-0261-4068} \and
Yisheng Guan\inst{1}\orcidlink{0000-0002-7011-0331} \and
Hong Zhang
\thanks{Corresponding author.}
\inst{2,3}\orcidlink{0000-0002-1677-6132}
}
\authorrunning{G. Chen et al.}
%
\institute{Guangdong
University of Technology, Guangzhou, China \email{2112001004@mail2.gdut.edu.cn, ysguan@gdut.edu.cn}
\and
Southern University of Science and Technology, Shenzhen, China
\email{\{hel,hzhang\}@sustech.edu.cn}\\
\and
Shenzhen Key Laboratory of Robotics and Computer Vision
\url{https://github.com/GCChen97/ppa4p3d}
}
\maketitle

\begin{abstract}
Current polarimetric 3D reconstruction methods, including those in the well-established shape from polarization literature, are all developed under the orthographic projection assumption.
In the case of a large field of view, however, this assumption does not hold and may result in significant reconstruction errors in methods that make this assumption.
To address this problem, we present the \emph{perspective phase angle (PPA) model} that is applicable to perspective cameras.
Compared with the orthographic model, the proposed PPA model accurately describes the relationship between polarization phase angle and surface normal under perspective projection.
In addition, the PPA model makes it possible to estimate surface normals from only one single-view phase angle map and does not suffer from the so-called $\pi$-ambiguity problem.
Experiments on real data show that the PPA model is more accurate for surface normal estimation with a perspective camera than the orthographic model.
\end{abstract}

\section{Introduction}

The property that the polarization state of light encodes geometric information of object surfaces has been researched in computer vision for decades. With the development of the division-of-focal-plane (DoFP) polarization image sensor \cite{polsensor} in recent years, there has been a resurgent interest in 3D reconstruction with polarization information.
In the last decade, it has been shown that polarization information can be used to enhance the performance of traditional reconstruction methods of textureless and non-Lambertian surfaces \cite{miyazaki2003polarization,atkinson2006recovery,shakeri2021icra,tozza2021uncalibrated}. For accurate reconstruction of objects, shape from polarization and photo-polarimetric stereo can recover fine-grain details \cite{tozza2021uncalibrated,fukao2021polarimetric} of the surfaces. For dense mapping in textureless or specular scenes, multi-view stereo can also be improved with polarimetric cues \cite{cui2017,yang2018,berger2017depth,shakeri2021icra}.
\begin{figure}[t]
	\centering
	\includegraphics[width=100mm]{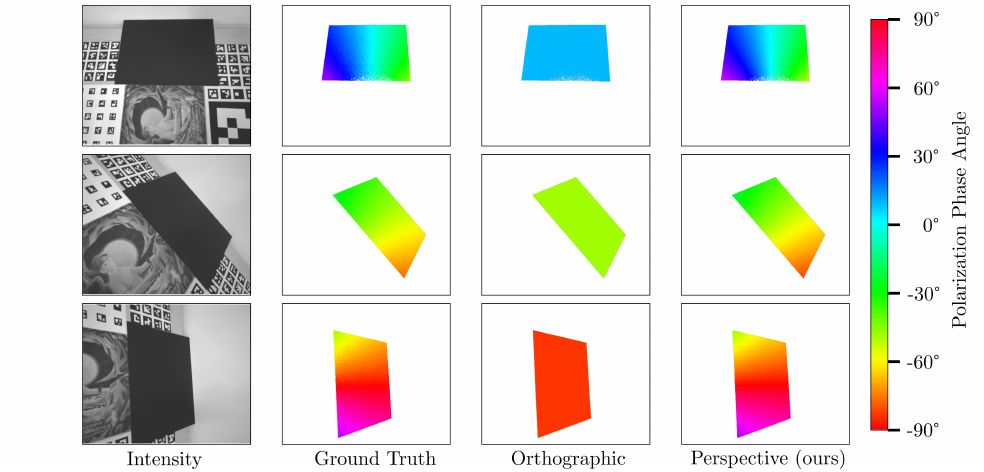}
	\caption{Comparison of polarization phase angle maps of a black glossy board in three views. The first column shows intensity images. The second column shows the ground-truth phase angle maps calculated from polarization images. The third and the fourth columns show the phase angle maps calculated from the OPA model and our proposed PPA model, respectively.}
	\label{fig:ex1}
\end{figure}

In computer vision and robotics, the use of perspective cameras is common. However, in all the previous literature on polarimetric 3D reconstruction, to the best of our knowledge, the use of polarization and its derivation are without exception under the orthographic projection assumption.
Therefore, the dense maps or shapes generated by these methods from polarization images captured by perspective cameras will be flawed without considering the perspective effect. In this paper, we present an accurate model of perspective cameras for polarimetric 3D reconstruction.

One of the key steps in polarimetric 3D reconstruction is the optimization of depth maps with a linear constraint on surface normals by the polarization phase angles.
In this paper, we refer to this constraint as the \emph{phase angle constraint}. This constraint was first proposed in \cite{wolff1990surface} under the orthographic projection assumption and has become a standard practice of utilizing the phase angles. In the literature, this constraint has been derived from a model, which we refer to as the \emph{orthographic phase angle (OPA) model} in this paper, where the azimuth of the normal is equivalent to the phase angle \cite{rahmann2001reconstruction,cui2017}. If observed objects are in the middle of the field of view in a 3D reconstruction application, orthographic projection could be a reasonable assumption. However, for dense mapping and large objects, this assumption is easily violated.

In this work, an alternative \emph{perspective phase angle (PPA) model} is developed to solve this problem. The PPA model is inspired by the geometric properties of a polarizer in \cite{korger2013polarizer}. Different from the OPA model, the proposed PPA model defines the phase angle as the direction of the intersecting line of the image plane and the plane of incident (PoI) spanned by the light ray and the surface normal (see Fig. \ref{fig:persp_model} for details). As shown in Fig. \ref{fig:ex1}, given the ground-truth normal of the black board in the images, the phase angle maps calculated using the proposed PPA model are much more accurate than those using the OPA model. Under perspective projection, we also derive a new linear constraint on the surface normal by the phase angle, which we refer to as the \emph{PPA constraint}.
As a useful by-product of our PPA model, the PPA constraint makes it possible to estimate surface normal using only one single-view phase angle map without suffering from the well-known $\pi$-ambiguity problem, which requires at least two views to resolve as shown in previous works \cite{wolff1990surface,rahmann2000polarization}.
In addition, the PPA constraint leads to improved the accuracy of normal estimation from phase angle maps of multiple views \cite{wolff1990surface,miyazaki2012pol}. The main contributions of this paper are summarized as follows:
\begin{itemize}
	\item A perspective phase angle model and a corresponding constraint on surface normals by polariztion phase angles are proposed. The model and the constraint serve as the basis for accurately estimating surface normals from phase angles.
	\item A novel method is developed to estimate surface normals from a \emph{single-view} phase angle map. The method does not suffer from the $\pi$-ambiguity problem as does a method using the orthographic phase angle model.
	\item We make use of the proposed model and the corresponding constraint to improve surface normal estimation from phase angle maps of \emph{multiple views}.
\end{itemize}

The rest of this paper is organized as follows. Section \ref{sec:related_works} overviews related works. Section \ref{sec:prelimi} reviews the utilization of polarization phase angles under orthographic projection. Section \ref{sec:proposed} describes our proposed PPA model, PPA constraint and normal estimation methods. Experimental evaluation is presented in Section \ref{sec:exps}. The conclusion of this paper and the discussions of our future work are presented in Section \ref{sec:final}.

\section{Related Works}\label{sec:related_works}
The proposed PPA model is fundamental to polarimetric 3D reconstruction and thus it is related to the following three topics: 1) shape and depth estimation from single-view polarization images, 2) multi-view reconstruction with additional polarization information and 3) camera pose estimation with polarization information.
\subsection{Polarimetric Single-view Shape and Depth Estimation}
This topic is closely related to two lines of research: shape from polarization (SfP) and photo-polarimetric stereo. SfP first estimates surface normals that are parameterized in the OPA model by azimuth and zenith angles from polarization phase angles and degree of linear polarization and then obtains Cartesian height maps by integrating the normal maps \cite{miyazaki2003polarization,atkinson2006recovery}. With additional spectral cues, refractive distortion \cite{kadambi2015polarized} of SfP can be solved \cite{huynh2010shape}. Alternatively, it has been shown that the surface normals, refractive indexes, and light directions can be estimated from photometric and polarimetric constraints \cite{ngo2015shape}. The recent work of DeepSfP \cite{ba2020deep} is the first attempt to use a convolutional neural network (CNN) to estimate normal maps from polarization images.
It is reasonable to expect more accurate surface normal estimation when these methods parameterize the surface normal with the help of the proposed PPA model in perspective cameras.
In fact, a recent work shows that learning-based SfP can benefit from considering the perspective effect \cite{Lei_2022_CVPR}.
Regardless, as a two-stage method, SfP is sensitive to noise.

As a one-stage method, photo-polarimetric stereo directly estimates a height map from polarization images and is able to avoid cumulative errors and suppress noise. By constructing linear constraints on surface heights from illumination and polarization, height map estimation is solved through the optimization of a non-convex cost function \cite{smith2016linear}. Yu \emph{et al.} \cite{yu2017shape} derives a fully differentiable method that is able to optimize a height map through non-linear least squares optimization. In \cite{tozza2021uncalibrated}, variations of the photo-polarimetric stereo method are unified as a framework incorporating different optional photometric and polarimetric constraints.
In these works, the OPA constraint is a key to exploiting polarization during height map estimation. In perspective cameras, the proposed PPA constraint can provide a more accurate description than the OPA constraint.

\subsection{Polarimetric Multi-view 3D Reconstruction}
Surface reconstruction can be solved by optimizing a set of functionals given phase angle maps of three different views \cite{rahmann2001reconstruction}. To address transparent objects, a two-view method \cite{miyazaki2003polarization} exploits phase angles and degree of linear polarization to solve correspondences and polarization ambiguity. Another two-view method \cite{atkinson2007shape} uses both polarimetric and photometric cues to estimate reflectance functions and reconstruct shapes of practical and complex objects. By combining space carving and normal estimation, \cite{miyazaki2012polarization} is able to solve polarization ambiguity problems and obtain more accurate reconstructions of black objects than pure space carving. 
Fukao \emph{et al.} \cite{fukao2021polarimetric} models polarized reflection of mesoscopic surfaces and proposes polarimetric normal stereo for estimating normals of mesoscopic surfaces whose polarization depends on illumination (i.e., polarization by light \cite{chen2018polarimetric}).
These works are all based on the OPA model without considering the perspective effect.

Recently, traditional multi-view 3D reconstruction methods enhanced by polarization prove to be able to densely reconstruct textureless and non-Lambertian surfaces under uncalibrated illumination \cite{cui2017,yang2018,shakeri2021icra}.
Cui \emph{et al.} \cite{cui2017} proposes polarimetric multi-view stereo to handle real-world objects with mixed polarization and solve polarization ambiguity problems. Yang \emph{et al.} \cite{yang2018} proposes a polarimetric monocular dense SLAM system that propagates sparse depths in textureless scenes in parallel. Shakeri \emph{et al.} \cite{shakeri2021icra} uses relative depth maps generated by a CNN to solve the $\pi/2$-ambiguity problem robustly and efficiently in polarimetric dense map reconstruction.
In these works, iso-depth contour tracing \cite{zhou2013multi} is a common step to propagate sparse depths on textureless or specular surfaces.
It is based on the proposition that, with the OPA model, the direction perpendicular to the phase angle is the tangent direction of an iso-depth contour \cite{cui2017}.
However, in a perspective camera, this proposition is only an approximation.

Besides, the OPA constraint can be integrated in stereo matching \cite{berger2017depth}, depth optimization \cite{yang2018} and mesh refinement \cite{zhao2020polarimetric}. However, only the first two components of the surface normal are involved in the constraint, in addition to its inaccuracy in perspective cameras. With the proposed PPA model, all three components of a surface normal are constrained and the constraint is theoretically accurate for a perspective camera.

\subsection{Polarimetric Camera Pose Estimation}
The polarization phase angle of light emitted from a surface point depends on the camera pose. Therefore, it is intuitive to use this cue for camera pose estimation. Chen \emph{et al.} \cite{chen2018polarimetric} proposes polarimetric three-view geometry that connects the phase angle and three-view geometry and theoretically requires six triplets of corresponding 2D points to determine the three rotations between the views. Different from using only phase angles in \cite{chen2018polarimetric}, Cui \emph{et al.} \cite{cui2019polarimetric} exploits full polarization information including degree of linear polarization to estimate relative poses so that only two 2D-point correspondences are needed. The method achieves competitive accuracy with the traditional five 2D-point method.
Since both works are developed without considering the perspective effect, the proposed PPA model can be used to generalize them to a perspective camera.

\section{Preliminaries}\label{sec:prelimi}
In this section, we review methods for phase angle estimation from polarization images, as well as the OPA model and the corresponding OPA constraint that are commonly adopted in the existing literature.

\subsection{Phase Angle Estimation}
As a function of the orientation of the polarizer $\phi$, the intensity of unpolarized light is attenuated sinusoidally as
$
	I(\phi) = I_{avg}+\rho I_{avg}\cos(2(\phi-\varphi))
$. The parameters of polarization state are the average intensity $I_{avg}$, the degree of linear polarization (DoLP) $\rho$ and the phase angle $\varphi$. The phase angle is also called the angle of linear polarization (AoLP) in \cite{zhao2020polarimetric,wu2020hdr,ting2021deep,shakeri2021icra}. In this paper, we uniformly use the term ``phase angle'' to refer to AoLP.

Given images captured through a polarizer at a minimum of three different orientations, the polarization state can be estimated by solving a linear system \cite{yang2018}. For a DoFP polarization camera that captures four images $I(0)$,\:$I(\frac{\pi}{4})$,\:$I(\frac{\pi}{2})$ and $I(\frac{3\pi}{4})$ in one shot, the polarization state can be extracted from the Stokes vector $\mathbf{s}=[s_0,s_1,s_2]^T$ as follows:
\begin{equation}\label{eq:phi_stokes}
	I_{avg}=\frac{s_0}{2},\:
	\varphi = \frac{1}{2}\text{arctan2}(s_2,s_1),\:
	\rho = \frac{\sqrt{s_1^2+s_2^2}}{s_0}
\end{equation}
where $s_o=I(0)+I(\frac{\pi}{2})$, $s_1=I(0)-I(\frac{\pi}{2})$ and $s_1=I(\frac{\pi}{4})-I(\frac{3\pi}{4})$.
Although it has been shown that further optimization of the polarization state from multi-channel polarization images is possible \cite{tozza2017linear}, for this paper, Eq. (\ref{eq:phi_stokes}) is sufficiently accurate for us to generate ground-truth phase angle maps from polarization images.

\subsection{The OPA Model}\label{sec:ortho_model}

The phase angles estimated from polarization images through Eq. (\ref{eq:phi_stokes}) are directly related to the surface normals off which light is reflected.
It is this property that is exploited in polarimetric 3D reconstruction research to estimate surface normal from polarization images. Model of projection, which defines how light enters the camera, is another critical consideration in order to establish the relationship between the phase angle and the surface normal.

The OPA model assumes that all light rays enter the camera in parallel, as shown by the blue plane in Fig. \ref{fig:persp_model}, so that the azimuth angle of the surface normal $\mathbf{n}$ is equivalent to the phase angle $\varphi_o$ up to a $\pi/2$-ambiguity and a $\pi$-ambiguity \cite{zhao2020polarimetric}. Specifically, $\mathbf{n}$ can be parameterized by the phase angle $\varphi_o$ and the zenith angle $\theta$ in the camera coordinate system:
\begin{equation}\label{eq:normal_param}
	\mathbf{n}=
	\left[
	\begin{array}{ccc}
		n_x,\;&
		n_y,\;&
		n_z
	\end{array}
	\right]^T
	=
	\left[
	\begin{array}{ccc}
		\cos\varphi_o\sin\theta,\;&
		-\sin\varphi_o\sin\theta,\;&
		\cos\theta
	\end{array}
	\right]^T
\end{equation}
With this model, $\varphi_o$ can be calculated from $\mathbf{n}$ as
\begin{equation}\label{eq:phi_ortho}
	\varphi_o=-\operatorname{arctan2}(n_y,\;n_x)
\end{equation}
We can use Eq. (\ref{eq:phi_ortho}) to evaluate the accuracy of the OPA model. I.e., if the OPA model is accurate, then the phase angle calculated by Eq. (\ref{eq:phi_ortho}) should agree with that estimated from the polarization images.
Although this model is based on the orthographic projection assumption, it is widely adopted in polarimetric 3D reconstruction methods even when polarization images are captured by a camera with perspective projection.
In addition, Eq. (\ref{eq:normal_param}) defines a constraint on $\mathbf{n}$ by $\varphi_o$, i.e., the OPA constraint:
\begin{equation}\label{eq:con_ortho}
	\left[
	\begin{array}{ccc}
		\sin\varphi_o,\; &
	 	\cos\varphi_o,\; & 0
	\end{array}
	\right]
	\cdot\mathbf{n}=0
\end{equation}
This linear constraint is commonly integrated in depth map optimization \cite{cui2017,yang2018,tozza2021uncalibrated}.
It is also the basis of iso-depth contour tracing \cite{alldrin2007toward,cui2017} as it gives the tangent direction of an iso-depth contour. Obviously, this constraint is strictly valid only in the case of orthographic projection.

\section{Phase Angle Model under Perspective Projection}\label{sec:proposed}
\begin{figure}[t]
	\centering
	\includegraphics[width=122mm]{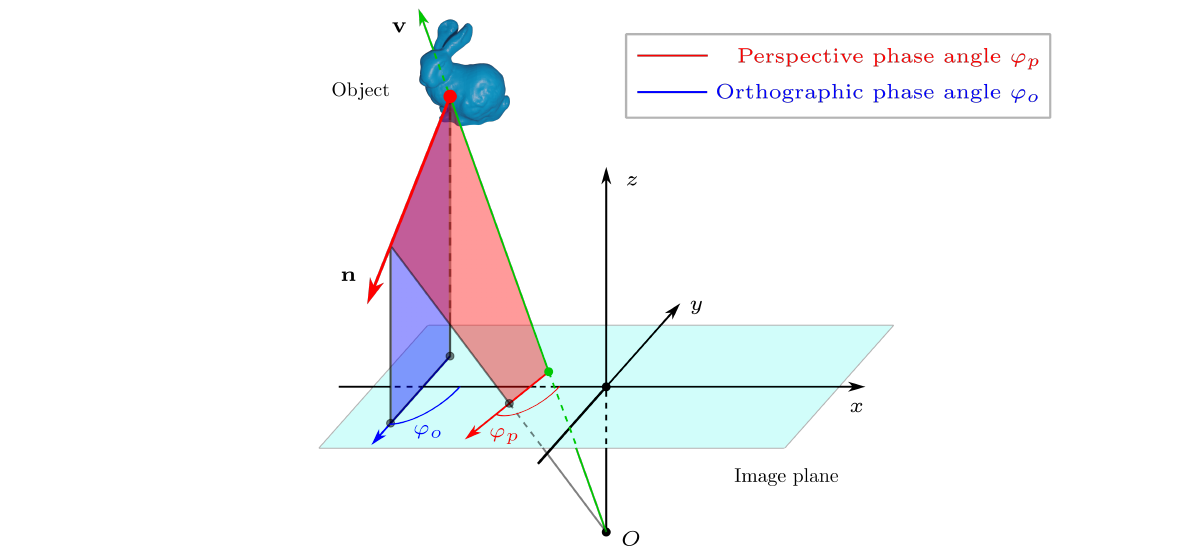}
	\caption{Definitions of the polarization phase angle of the two models under orthographic projection ($\varphi_o$) and perspective projection ($\varphi_p$). $O$ is the optical center of the camera. The blue plane and the red plane are the two PoIs, corresponding to the two (orthographic and perspective) projection models.}
	\label{fig:persp_model}
\end{figure}

In the perspective camera model, light rays that enter the camera are subject to the perspective effect as illustrated by the red plane in Fig. \ref{fig:persp_model}. As a result, the estimated phase angle by Eq. (\ref{eq:phi_stokes}) not only depends on the direction of the surface normal $\mathbf{n}$ but also on the direction $\mathbf{v}$ in which the light ray enters the camera.
In this section, the PPA model is developed to describe the relationship between the polarization phase angle and the surface normal. The PPA constraint naturally results from the PPA model as well as two methods for normal estimation from the phase angles.
\subsection{The PPA Model}\label{sec:persp_model}

The geometric properties of a polarizer in \cite{korger2013polarizer} inspire us to model the phase angle as the direction of the intersecting line of the image plane and the PoI spanned by the light ray and the surface normal. As shown in Fig. \ref{fig:persp_model}, there is an obvious difference between the two definitions of the phase angle, depending upon the assumed type of projection. 

Specifically, let the optical axis be $\mathbf{z}=[0,0,1]^T$ and the light ray be $\mathbf{v}=[v_x,v_y,v_z]^T=K^{-1}\mathbf{x}/\|K^{-1}\mathbf{x}\|$, for an image point at the pixel coordinates $\mathbf{x}=[u,v,1]^T$ and a camera with intrinsic matrix $K$. From Fig. \ref{fig:persp_model}, the PPA model can be formulated as follows:
\begin{equation}
	\begin{aligned}\label{eq:model}
		\mathbf{z}\times (\mathbf{v}\times \mathbf{n})
		=
		\left[
		\begin{array}{ccc}
			-v_zn_x+v_xn_z,\; & -v_zn_y+v_yn_z,\; & 0
		\end{array}
		\right]^T
		=c\cdot\mathbf{d}
	\end{aligned}
\end{equation}
where $\mathbf{z}$, $\mathbf{v}$ and $\mathbf{n}$ are normalized to 1, $c$ is a constant and $\mathbf{d}=[\cos\varphi_p,-\sin\varphi_p,0]^T$.
From Eq. (\ref{eq:model}), the phase angle $\varphi_p$ can be obtained from $\mathbf{n}$ and $\mathbf{v}$ as:
\begin{equation}\label{eq:phi_persp}
	\varphi_p = -\operatorname{\arctan2}(-v_zn_y+v_yn_z,\,-v_zn_x+v_xn_z)
\end{equation}
Similar to Eq. (\ref{eq:phi_ortho}), Eq. (\ref{eq:phi_persp}) can also be used for evaluating the accuracy of the PPA model.
Different from $\varphi_o$, $\varphi_p$ not only depends on the surface normal $\mathbf{n}$ but also on the light ray $\mathbf{v}$. 

In addition, $\mathbf{n}$ can be parameterized by $\varphi_p$ and $\theta$ as
$
\mathbf{n}= -(e^{\theta \mathbf{a}^{\land}_p}) \mathbf{v}
$
where $\mathbf{a}_p=\mathbf{d}\times \mathbf{v}/\|\mathbf{d}\times \mathbf{v}\|$ is the normal of the PoI and $\mathbf{r}^{\land}$ represents the skew-symmetric matrix of $\mathbf{r} \in \mathbb{R}^3$. Eq. (\ref{eq:normal_param}) can also be reformulated into a similar form as
$
	\mathbf{n}= -(e^{\theta \mathbf{a}^{\land}_o}) \mathbf{v}
$
where $\mathbf{a}_o=[-\sin\varphi_o,\cos\varphi_o,0]^T$. Obviously, the difference of the two parameterizations of $\mathbf{n}$ is the axis about which $\mathbf{v}$ rotates.

With the PPA model, the PPA constraint can be derived from Eq. (\ref{eq:model}) as follows:
\begin{equation}\label{eq:con_persp}
	\left[
	\begin{array}{ccc}
		\sin\varphi_p,\; & \cos\varphi_p,\; & -\displaystyle\frac{(v_y\cos\varphi_p+v_x\sin\varphi_p)}{v_z}
	\end{array}
	\right]
	\cdot\mathbf{n}=0
\end{equation}
Note that Eq. (\ref{eq:con_persp}) can be used just as the OPA constraint Eq. (\ref{eq:con_ortho}), without the knowledge of object geometry since $\mathbf{v}$ is a scene-independent directional vector.
Compared with Eq. (\ref{eq:con_ortho}), Eq. (\ref{eq:con_persp}) has an additional constraint on the third component of the normal.
This additional constraint is critical in allowing us to estimate $\mathbf{n}$ of a planar surface from a single view as we will show in the next section.
Table \ref{table:two_models} summaries the definitions of the phase angle, the normal and the constraint of the OPA and PPA models.
\begin{table}
	\caption{Summary of the OPA and PPA models}
	\label{table:two_models}
	\begin{center}
		\renewcommand{\arraystretch}{1.7}

		\begin{tabular}{c|c|c}
			\hline
			 & OPA model & PPA model\\
			\hline
			Phase angle
			&
			$-\operatorname{arctan2}(n_y,\;n_x)$
			&
			$\begin{aligned}-\operatorname{\arctan2}(&-v_zn_y+v_yn_z,\\&-v_zn_x+v_xn_z)\end{aligned}$
			\\
			Normal
			& 
			$\mathbf{n}=
			-e^{\theta \mathbf{a}^{\land}_o}\cdot \mathbf{v}$
			&
			$\mathbf{n}=-e^{\theta \mathbf{a}^{\land}_p }\cdot \mathbf{v}$
			\\
			Constraint
			&
			$\begin{aligned}
				&\left[\begin{array}{c}
					\sin\varphi_o\\
					\cos\varphi_o\\
					0
				\end{array}\right]^{T}
				\cdot\mathbf{n}
				=0\end{aligned}$
			&
			$\begin{aligned}
				&\left[
				\begin{array}{c}
					\sin\varphi_p \\ \cos\varphi_p \\ -\displaystyle\frac{(v_y\cos\varphi_p+v_x\sin\varphi_p)}{v_z}
				\end{array}
				\right]^{T}
				\cdot\mathbf{n}
				=0\end{aligned}$
			\\
			\hline
		\end{tabular}
	\end{center}
\end{table}

\subsection{Relations with the OPA Model}\label{sec:relation}

Comparing the two constraints in Table \ref{table:two_models}, the coefficient $-(v_y\cos\varphi_p+v_x\sin\varphi_p)/v_z$ implies that there exist four cases in which the two models are equivalent:
\begin{enumerate}
	\setlength{\itemsep}{5pt}
	\item $\mathbf{v}=[0,0,1]^T$, when the light ray is parallel to the optical axis, which corresponds to orthographic projection.
	\item $[v_x,v_y]^T\parallel [\cos\varphi_p,-\sin\varphi_p]^T \parallel [n_x,n_y]^T$, when the PoI is perpendicular to the image plane.
	\item $n_z\to 1$, when the normal tends to be parallel to the optical axis. In this case, the PoI is also perpendicular to the image plane.
	\item $n_z=0$, when the normal is perpendicular to the optical axis.
\end{enumerate}
Besides, according to Eq. (\ref{eq:phi_ortho}) and Eq. (\ref{eq:con_ortho}), if $n_z=1$, the OPA model and the OPA constraint will become degenerate whereas ours will still be useful.

\begin{figure}[t] 
	\centering
	\subfloat[]{\includegraphics[height=1.2in]{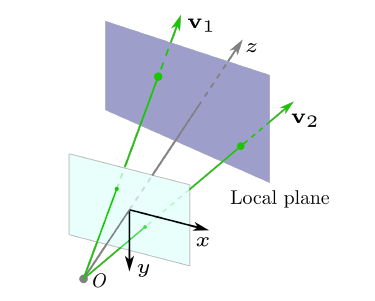}}
	\subfloat[]{\includegraphics[height=1.2in]{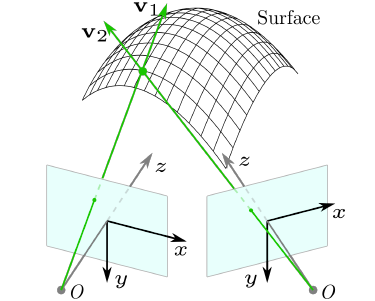}}
	\caption{Two cases of surface normal estimation from phase angle map(s). (a) Single-view normal estimation and (b) Multi-view normal estimation.}
	\label{fig:two_methos}
\end{figure}

\subsection{Normal Estimation}\label{sec:normal}
Theoretically, a surface normal can be solved from at least its two observations with the PPA constraint. The observations can either be the phase angles from multiple pixels in the same phase angle map or from phase angle maps of multiple views as shown in Fig. \ref{fig:two_methos}. The two cases lead to the following two methods.

\subsubsection{Single-view Normal Estimation.}\label{sec:normal_single_view}
In the same phase angle map, if a set of phase angles of the points share the same surface normal, e.g., points in a local plane or points in different parallel planes, they can be used for estimating the normal. Let $\mathbf{m}$ denote the coefficients in Eq. (\ref{eq:con_persp}) as $\mathbf{m}=[\sin\varphi_p,\cos\varphi_p,-(v_y\cos\varphi_p\\+v_x\sin\varphi_p)/v_z]^T$. Given $P$ points that have the same normal $\mathbf{n}$, a coefficient matrix $\mathbf{M}_1$ can be constructed from their coefficients:
\begin{equation}\label{eq:coeff}
	\mathbf{M}_1 = 
	\left[
	\begin{array}{ccc}
		\mathbf{m}_1^T,\;
		\cdots,\;
		\mathbf{m}_P^T
	\end{array}
	\right]^T
\end{equation}
then $\mathbf{n}$ can be obtained by the eigen decomposition of $\mathbf{M}_1^T\mathbf{M}_1$. This method directly solves $\mathbf{n}$ with only one single-view phase angle map and does not suffer from the $\pi$-ambiguity problem, while previous works based on the OPA model require at least two views \cite{wolff1990surface,rahmann2000polarization}.
Ideally the rank of $\mathbf{M}_1$ should be exactly two so that $\mathbf{n}$ can be solved.
In practice, with image noise, we can construct a well-conditioned $\mathbf{M}_1$ with more than two points that are on the same plane for solving $\mathbf{n}$ as we will show in Section \ref{sec:exp_normal}.

\subsubsection{Multi-view Normal Estimation.}\label{sec:normal_multiview}
If a point is observed in $K$ views, another coefficients matrix $\mathbf{M}_2$ can be constructed as follows:
\begin{equation}\label{eq:coeff_2}
	\mathbf{M}_2 = 
	\left[
	\begin{array}{ccc}
		\mathbf{m}_1^T \mathbf{R}_1,\;
		\cdots,\;
		\mathbf{m}_K^T \mathbf{R}_K
	\end{array}
	\right]^T
\end{equation}
where $\mathbf{R}_k$ ($k=1,2,\cdots,K$) is a camera rotation matrix.
The normal $\mathbf{n}$ can be solved as in the case of single-view normal estimation above. Similar to $\mathbf{M}_1$, $\mathbf{n}$ can be solved if the rank of $\mathbf{M}_2$ is two and well conditioned. Therefore, observations made from sufficiently different camera poses are desirable. 
This method is the generalization of the method proposed in \cite{wolff1990surface,miyazaki2012pol} to the case of perspective projection. 
We will verify that a method using the PPA constraint is more accurate in a perspective camera than using the OPA constraint in Section \ref{sec:exp_normal}.

\section{Experimental Evaluation}\label{sec:exps}

In this section, we present the experimental results that verify the proposed PPA model and PPA constraint in a perspective camera. Details on the experimental dataset and experimental settings are provided in Section \ref{sec:settings}. In Section \ref{sec:exp_acc}, we evaluate the accuracy of the PPA model and analyze its phase angle estimation error. To verify that the PPA model is beneficial to polarimetric 3D reconstruction, we conduct experiments on normal estimation in Section \ref{sec:exp_normal} and on contour tracing in Section \ref{sec:exp_contour}.

\begin{figure}[t]
	\centering
	\includegraphics[width=122mm]{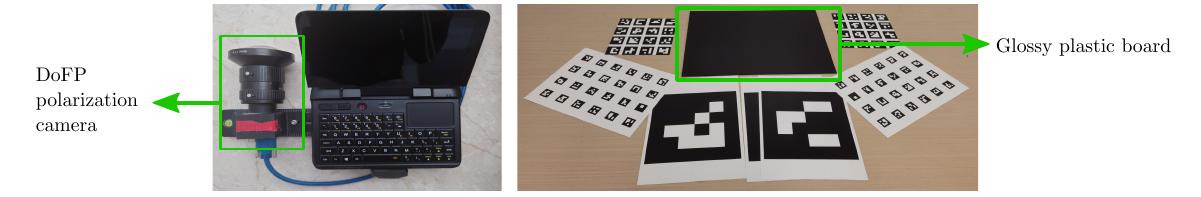}
	\caption{The capture setup of polarization images of the dataset. The glossy and black plastic board is captured by the DoFP polarization camera at random view points. The AR tags are used for estimating the ground-truth poses of the camera and the board.}
	\label{fig:setup}
\end{figure}
\subsection{Dataset	and Experimental Settings}\label{sec:settings}
An image capture setup is shown in Fig. \ref{fig:setup}.
Our camera has a lens with a focal length of 6 mm and a field of view of approximately $86.6^\circ$. Perspective projection is therefore appropriate to model its geometry. We perform camera calibration first to obtain its intrinsics, and use the distortion coefficients to undistort the images before they are used in our experiments. The calibration matrix $K$ is used to generate the light ray $\mathbf{v}$ in homogeneous coordinates.

We capture a glossy and black plastic board with a size of 300 mm by 400 mm on a table by a handheld DoFP polarization camera \cite{polcam}.
The setting is such that:
1) the board is specular-reflection-dominant so that the $\pi/2$ ambiguity problem can be easily solved
and 2) the phase angle maps of the board can be reasonably estimated.
Although this setting could be perceived as being limited, it is carefully chosen to verify the proposed model accurately and conveniently.

We create a dataset that contains $282$ groups of grayscale polarization images (four images per group) of the board captured at random view points.
The camera poses and the ground-truth normal of the board are estimated with the help of AR tags placed on the table. Ground-truth phase angle maps are calculated from polarization images by Eq. (\ref{eq:phi_stokes}).
To reduce the influence of noise, we only use the pixels with DoLP higher than $0.1$ in the region of the board, and apply Gaussian blur to the images before calculating phase angle maps and DoLP maps. This dataset is used in all the following experiments.


\begin{figure*}[t]
	\centering
	\includegraphics[width=122mm]{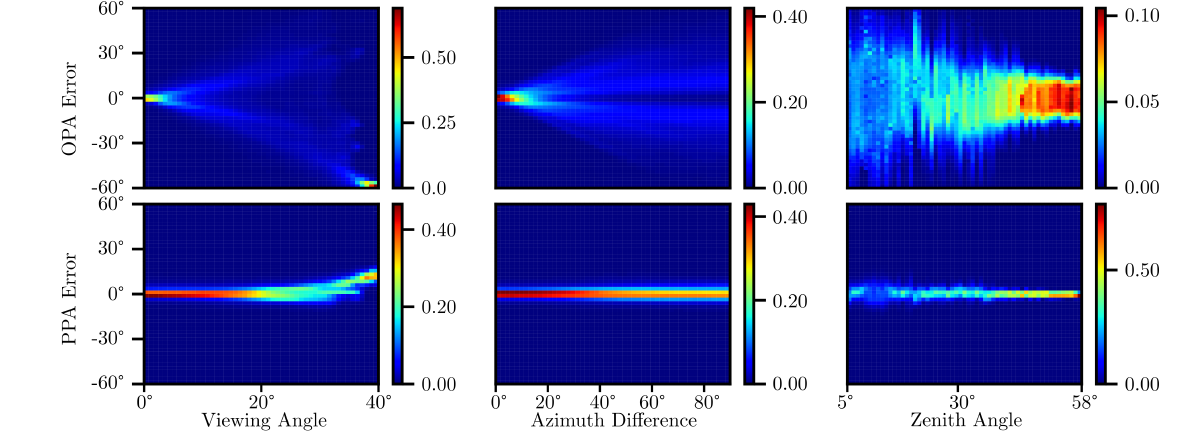}
	\caption{The phase angle error distributions of the two models. The first row shows the distributions of the OPA model and the second row corresponds to the PPA model. Each column in one subfigure represents the one-dimension error distribution of a specific angle. The density is expressed in the form of pseudo color.}
	\label{fig:residuals}
\end{figure*}

\subsection{Accuracy of the PPA Model}\label{sec:exp_acc}
To evaluate the accuracy of the OPA model and the PPA model,
we calculate the phase angle of every pixel with Eq. (\ref{eq:phi_ortho}) and Eq. (\ref{eq:phi_persp}), respectively, given the ground-truth normals, and compare the results to the estimated phase angle by Eq. (\ref{eq:phi_stokes}).
The mean and the root mean square error (RMSE) of the phase angle errors of the OPA model are $-0.18^{\circ}$, and $20.90^{\circ}$ respectively, and the ones of our proposed PPA model are $0.35^{\circ}$ and $5.90^{\circ}$ respectively.
Although the phase angle error of both the OPA and the PPA models is unbiased with a mean that is close to zero, the RMSE of the phase angle estimated by the PPA model is only $28\%$ of that by the OPA model in relative terms, and small in absolute terms from a practical point of view.
As shown in Fig. \ref{fig:ex1}, the PPA model accurately describes the spatial variation of the phase angle while the phase angle maps of the OPA model are spatially uniform and inaccurate, in comparison with the ground truth.

In addition, we plot the error distributions with respect to the viewing angle (the angle between the light ray and the normal), the azimuth difference (the angle between the azimuths of the light ray and the normal) and the zenith angle of the normal.
As shown in Fig. \ref{fig:residuals}, the deviations of the PPA error are much smaller than those of the OPA error.
It is shown that pixels closer to the edges of the views have larger errors with the OPA model.
Additionally, the errors are all close to zeros in the first three equivalent cases stated in Section \ref{sec:relation}.
The high density at viewing angles near $40^{\circ}$ in the first column of Fig. \ref{fig:residuals} is likely a result of the nonuniform distribution of the positions of the board in the images since they correspond to the pixels at the edges of the views.

\subsection{Accuracy of Normal Estimation}\label{sec:exp_normal}
\subsubsection{Single-view Normal Estimation.}\label{sec:single_view_exp}
As has been mentioned, it is possible to estimate surface normal from a single view by a polarization camera. Methods based on the OPA model however suffer from the $\pi$-ambiguity problem.
In contrast, given a single-view phase angle map, surface normal estimation can be solved with the proposed PPA constraint without suffering from the $\pi$-ambiguity problem.
Therefore, we only evaluate our method developed in Section \ref{sec:normal_single_view} with the PPA constraint for the normal of a planar surface.
\begin{figure}[t]
	\centering
	\includegraphics[width=100mm]{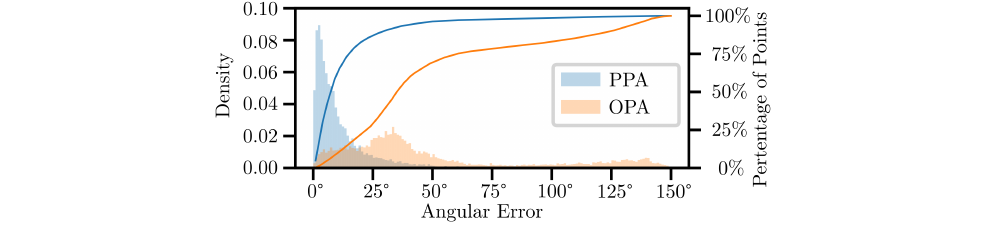}
	\caption{Histogram and cumulative distribution function curves of angular errors of three-view normal estimation.}
	\label{fig:normal}
\end{figure}
\begin{figure*}[t]
	\centering
	\subfloat{\includegraphics[width=1.601in]{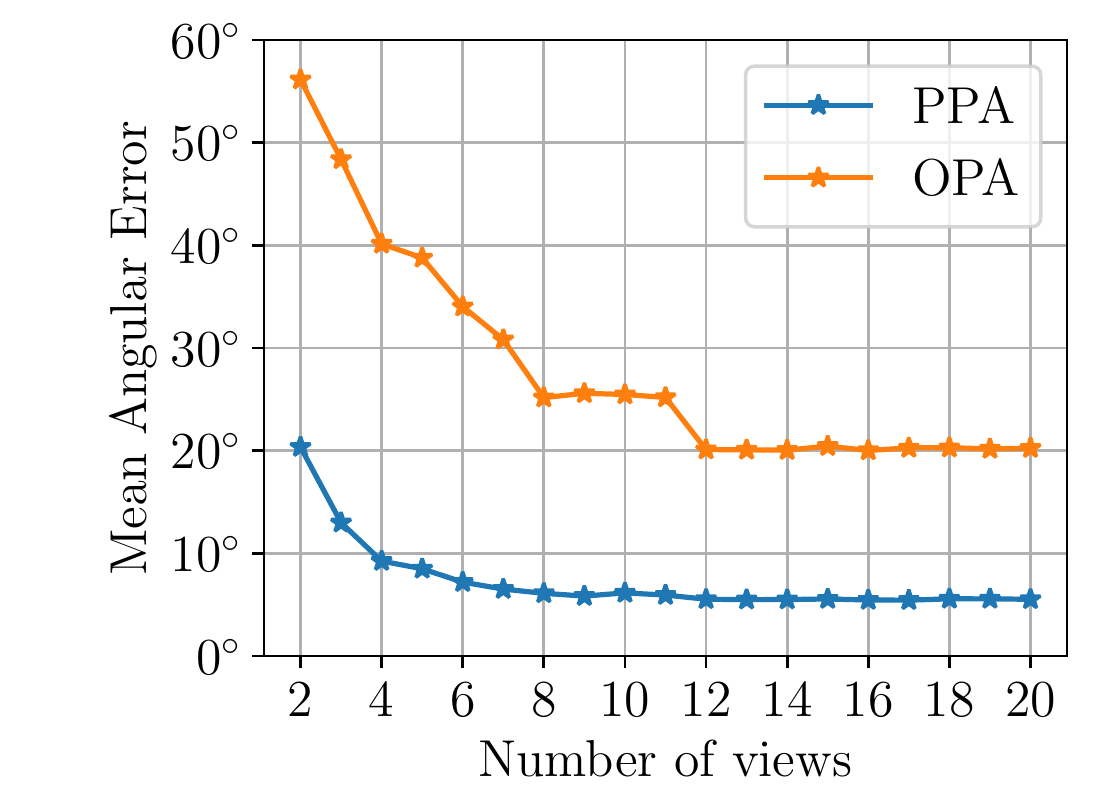}}
	\subfloat{\includegraphics[width=1.601in]{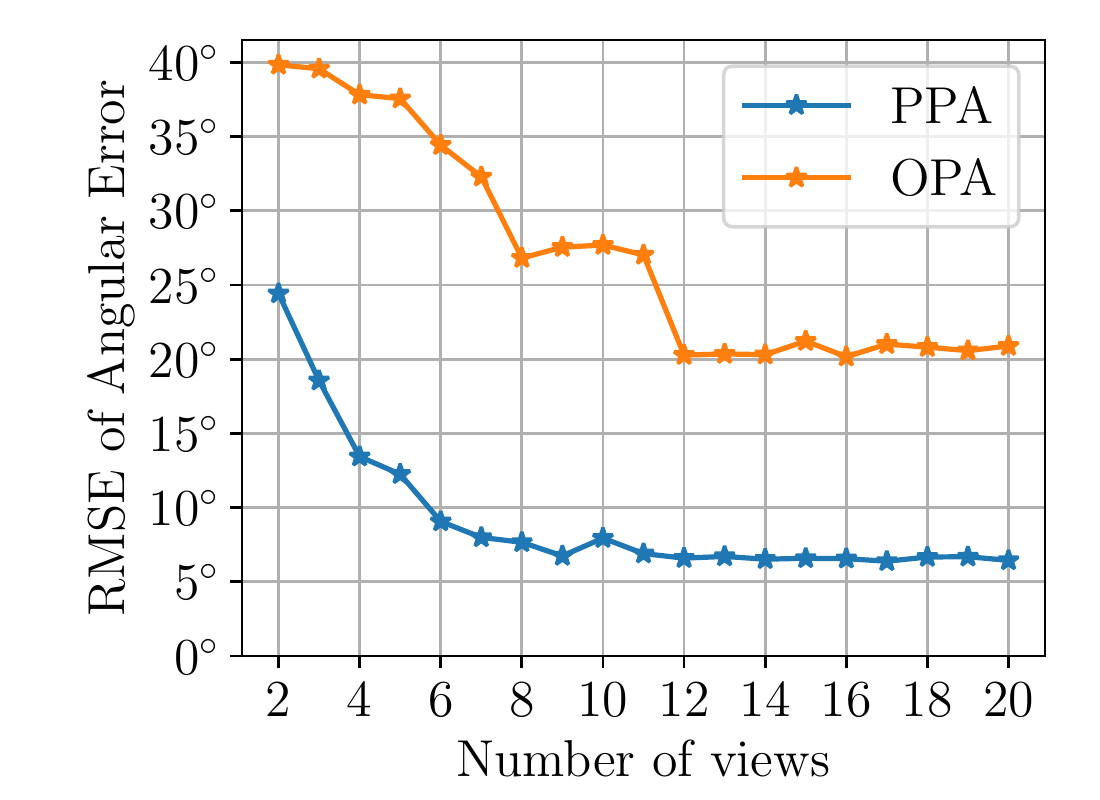}}
	\subfloat{\includegraphics[width=1.601in]{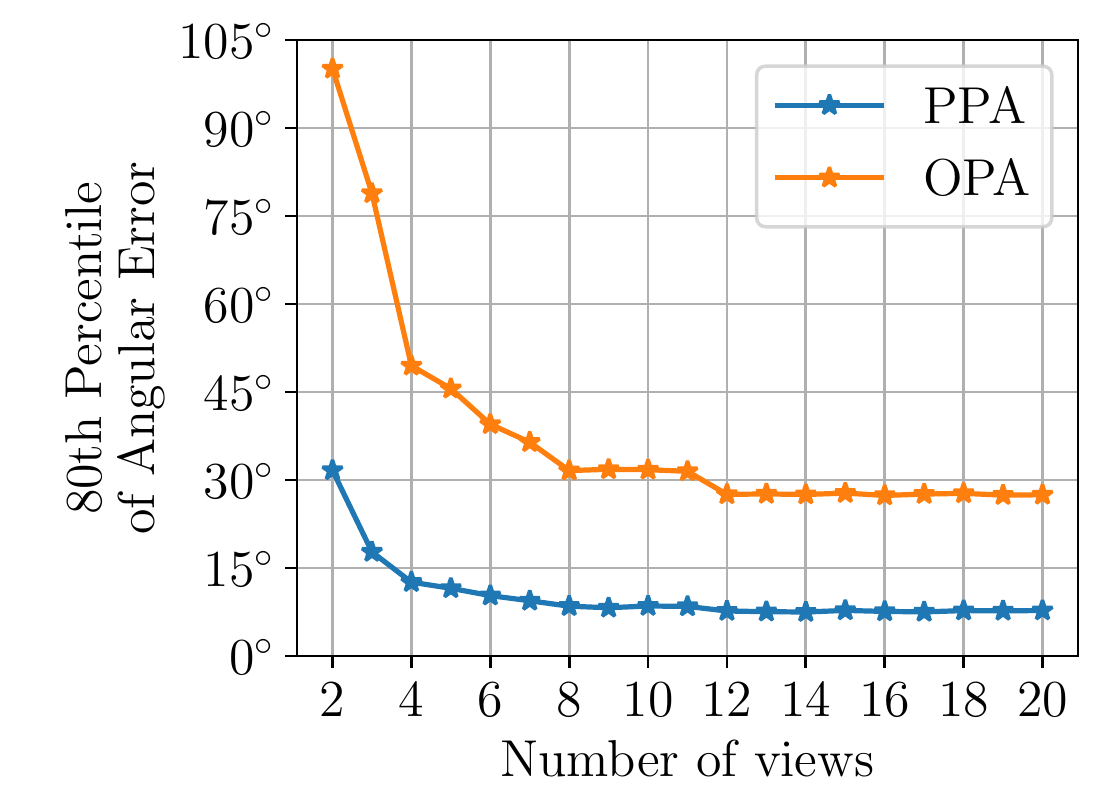}}
	\caption{Angular error with the increase of the number of views.}
	\label{fig:residuals_2}
\end{figure*}

In this experiment, for every image in the dataset introduced in Section \ref{sec:settings}, we use the phase angles of the pixels in the region of the black board in Fig. \ref{fig:setup} to estimate its normal.
The number of the pixels contributing to the normal in one estimation varies from 100,000 to 300,000, depending on the size of area of the board in the image. We obtain 282 estimated normals among the 282 images (see the estimated normals in our supplementary video).
The mean and the RMSE of the angular errors of these estimated normals are $2.68^{\circ}$ and $1.16^{\circ}$, respectively.
Such excellent performance is in part due to our highly accurate PPA model in describing the image formation process and in part due to the simplicity of the scene (planar surface) and a large number of measurements available.

\subsubsection{Multi-view Normal Estimation.}
As well, it is possible to estimate surface normal from multiple views of a polarization camera.
To establish the superiority of our proposed model, 
we compare the accuracy of multi-view normal estimation with the proposed method using the PPA constraint developed in Section \ref{sec:normal_multiview} and the one using the OPA constraint proposed in \cite{wolff1990surface,rahmann2000polarization}, respectively.

In this experiment, we randomly sample $10,000$ 3D points on the board among the 282 images and individually estimate the normals of the points from multiple views. For every point, the number of views for its normal estimation varies from two to 20, and the views are selected using the code of ACMM \cite{ACMM} and the ground-truth poses of the camera and the board are used to resolve the pixel correspondences among the multiple views.
\begin{figure}[t]
	\centering
	\subfloat[]{\includegraphics[width=1.3in]{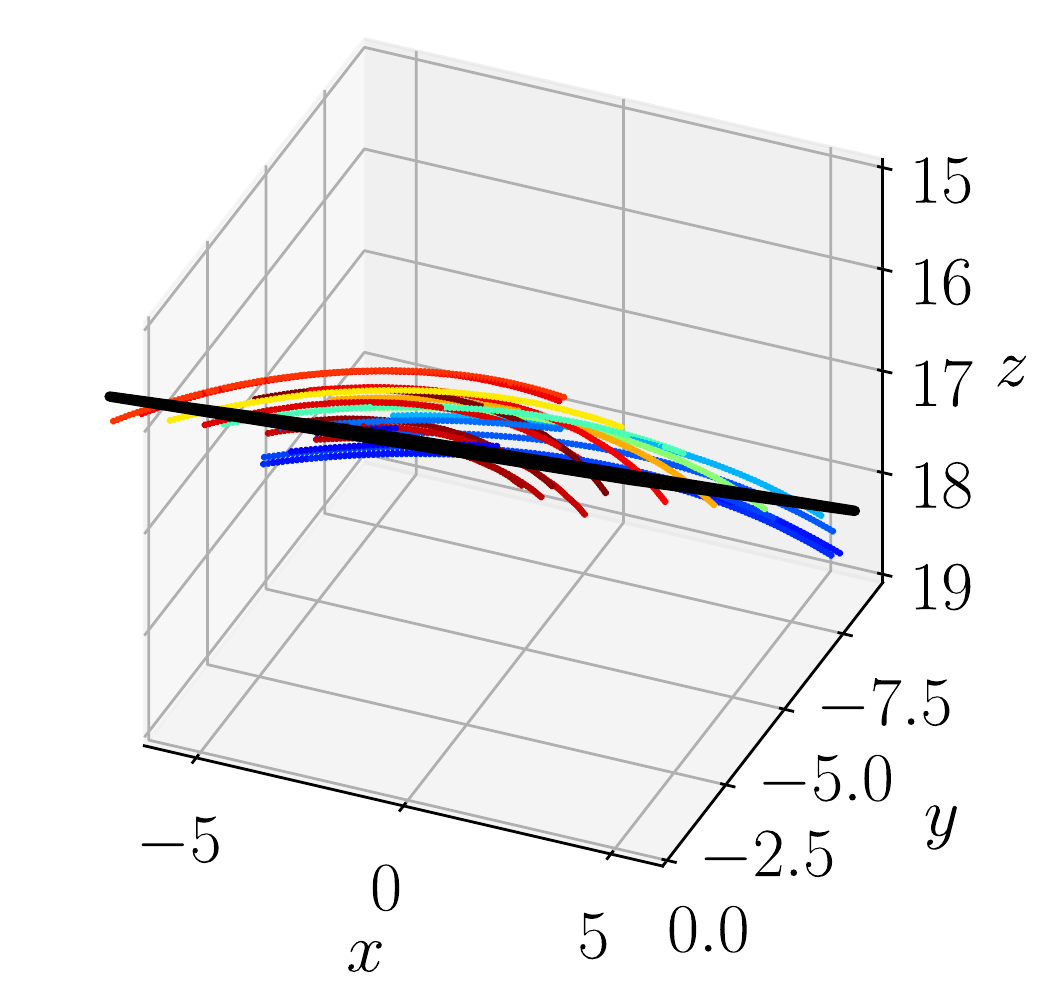}}
	\subfloat[]{\includegraphics[width=1.3in]{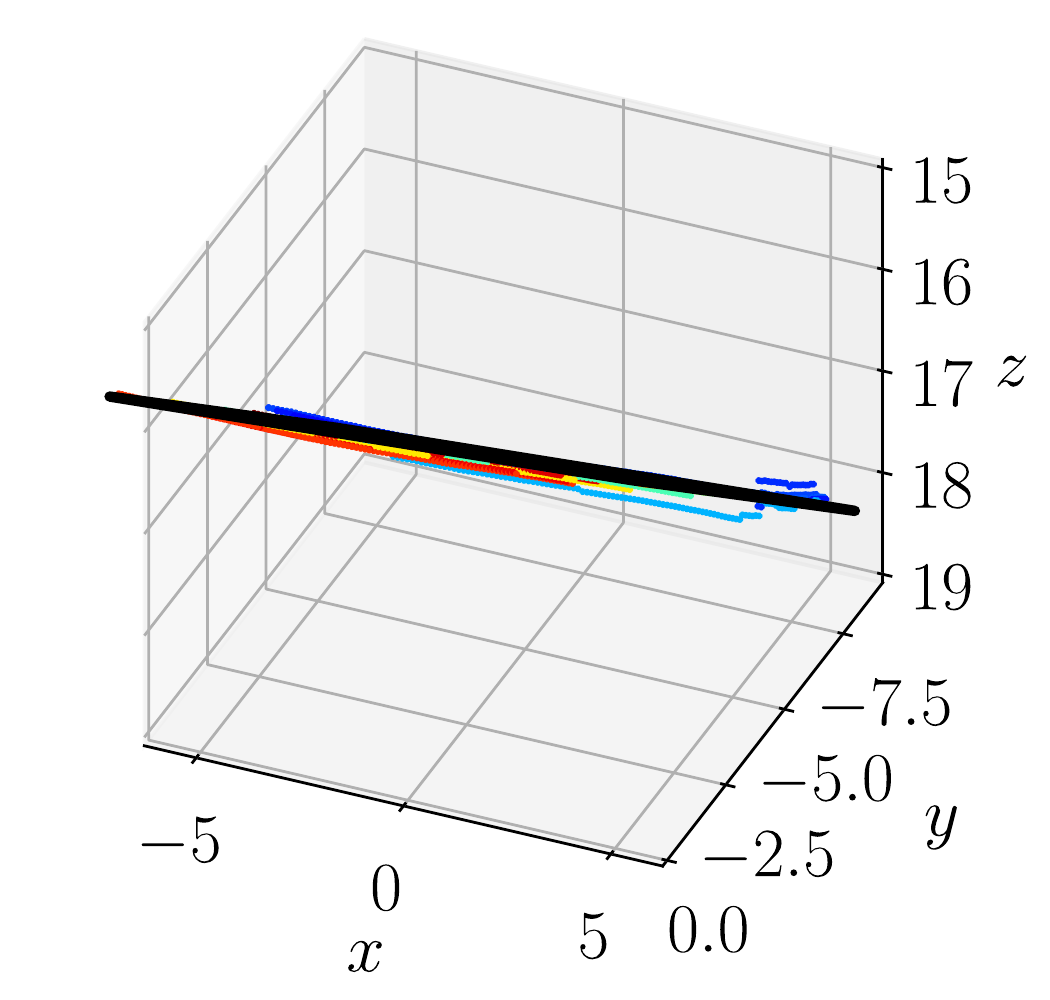}}
	\caption{Contours of a plane generated by the iso-depth contour tracing and our method respectively. (a) Contours from iso-depth tracing using the OPA model and (b) Contours using our PPA model. The viewing direction is parallel to the plane and the black line represents the plane.}
	\label{fig:contour}
\end{figure}

Fig. \ref{fig:normal} shows the distribution of the angular errors of the three-view case. $80\%$ of the results of our method have errors smaller than $25^{\circ}$ compared with only $26\%$ of the one using the OPA constraint.
Fig. \ref{fig:residuals_2} shows the normal estimation errors with different number of views.
Although the accuracy of these two methods can be improved with the increase of the number of views, the errors of our method with three views are already $50\%$ smaller than those using the OPA model with more than ten views.
The results of this experiment can be visualized in our supplementary video.
Note that the error of the estimated normal in this case is significantly larger than that in the single-view case, mostly because of the small number of measurements (2-20) used in estimating each normal and the uncertainty in the relative camera poses used to solve data association. 
Nonetheless, the superiority of the PPA model over the OPA model is clearly established.



\subsection{Comparison of Contour Tracing}\label{sec:exp_contour}

This experiment is designed to illustrate the influence of the perspective effect on the iso-depth surface contour tracing with the OPA model and to show the accuracy of the proposed PPA model.
We sample 20 seed 3D points on the edge of the board and propagate their depths to generate contours of the board. With the OPA model, the iso-depth contour tracing requires only the depths and a single-view phase angle map. However, according to the proposed PPA model, the iso-depth contour tracing is infeasible in perspective cameras. Therefore, we estimate the normals of the points through the method developed in Section \ref{sec:normal_multiview} with two views for generating surface contours. To clearly compare the quality of the contours, our method is set to generate 3D contours that have the same 2D projections on the image plane as those by iso-depth contour tracing. The propagation step size is set to $0.5$ pixel, which is the same as that in \cite{cui2017}.

The iso-depth contours of the board are expected to be straight in perspective cameras. However, as shown in Fig. \ref{fig:contour}, the contours generated by iso-depth contour tracing using the OPA model are all curved and lie out of the ground-truth plane, while those generated by our PPA model are well aligned with the ground-truth plane.
We also compute the RMSE of the distances between the points on the contours and the plane. The RMSE of the contours using our PPA model is $2.2$ mm, $25\%$ of that using the OPA model which is $9.6$ mm.

\section{Conclusions and Future Work}\label{sec:final}
In this paper, we present the perspective phase angle (PPA) model as a superior alternative to the orthographic phase angle (OPA) model for accurately utilizing polarization phase angles in 3D reconstruction with perspective polarization cameras.
The PPA model defines the polarization phase angle as the direction of the intersecting line of the image plane and the plane of incident, and hence allows the perspective effect to be considered in estimating surface normals from the phase angles and in defining the constraint on surface normal by the phase angle.
In addition, a novel method for surface normal estimation from a single-view phase angle map naturally results from the PPA model that does not suffer from the well-known $\pi$-ambiguity problem as in the traditional orthographic model.
Experimental results on real data validate that our PPA model is more accurate than the commonly adopted OPA model in perspective cameras.
Overall, we demonstrate the necessity of considering the perspective effect in polarimetric 3D reconstruction and propose the PPA model for realizing it.

As a limitation of our work, we have so far only conducted experiments on surface normal estimation and contour tracing. We have not used our model in solving other problems related to polarimetric 3D reconstruction.
Our immediate future plan includes improving polarimetric 3D reconstruction methods with the proposed PPA model.
We are also interested in synthesizing polarization images with open-source datasets and the PPA model for data-driven approaches.
We leave the above interesting problems as our future research.
\\
\\
\noindent\textbf{Acknowledgments.}
We thank the reviewers for their valuable feedback.
This work was done while Guangcheng Chen was a visiting student at Southern University of Science and Technology.
This work was supported in part by the National Natural Science Foundation of China under Grant No. 62173096,
in part by the Leading Talents Program of Guangdong Province under Grant No. 2016LJ06G498 and 2019QN01X761,
in part by Guangdong Province Special Fund for Modern Agricultural Industry Common Key Technology R\&D Innovation Team under Grant No. 2019KJ129,
in part by Guangdong Yangfan Program for Innovative and Entrepreneurial Teams under Grant No. 2017YT05G026.

\clearpage
%
%
\bibliographystyle{splncs04}
\bibliography{ref}

\begin{thebibliography}{10}
\providecommand{\url}[1]{\texttt{#1}}
\providecommand{\urlprefix}{URL }
\providecommand{\doi}[1]{https://doi.org/#1}

\bibitem{alldrin2007toward}
Alldrin, N.G., Kriegman, D.J.: Toward reconstructing surfaces with arbitrary
  isotropic reflectance: A stratified photometric stereo approach. In: 2007
  IEEE 11th International Conference on Computer Vision. pp.~1--8. IEEE (2007)

\bibitem{atkinson2006recovery}
Atkinson, G.A., Hancock, E.R.: Recovery of surface orientation from diffuse
  polarization. IEEE transactions on image processing  \textbf{15}(6),
  1653--1664 (2006)

\bibitem{atkinson2007shape}
Atkinson, G.A., Hancock, E.R.: Shape estimation using polarization and shading
  from two views. IEEE transactions on pattern analysis and machine
  intelligence  \textbf{29}(11),  2001--2017 (2007)

\bibitem{ba2020deep}
Ba, Y., Gilbert, A., Wang, F., Yang, J., Chen, R., Wang, Y., Yan, L., Shi, B.,
  Kadambi, A.: Deep shape from polarization. In: European Conference on
  Computer Vision. pp. 554--571. Springer (2020)

\bibitem{berger2017depth}
Berger, K., Voorhies, R., Matthies, L.H.: Depth from stereo polarization in
  specular scenes for urban robotics. In: 2017 IEEE international conference on
  robotics and automation (ICRA). pp. 1966--1973. IEEE (2017)

\bibitem{chen2018polarimetric}
Chen, L., Zheng, Y., Subpa-Asa, A., Sato, I.: Polarimetric three-view geometry.
  In: Proceedings of the European Conference on Computer Vision (ECCV). pp.
  20--36 (2018)

\bibitem{cui2017}
Cui, Z., Gu, J., Shi, B., Tan, P., Kautz, J.: Polarimetric multi-view stereo.
  In: Proceedings of the IEEE conference on computer vision and pattern
  recognition. pp. 1558--1567 (2017)

\bibitem{cui2019polarimetric}
Cui, Z., Larsson, V., Pollefeys, M.: Polarimetric relative pose estimation. In:
  Proceedings of the IEEE/CVF International Conference on Computer Vision. pp.
  2671--2680 (2019)

\bibitem{polcam}
FLIR: Blackfly s usb3,
  \url{https://www.flir.com.au/products/blackfly-s-usb3/?model=BFS-U3-51S5P-C},
  accessed 2022-02-14

\bibitem{fukao2021polarimetric}
Fukao, Y., Kawahara, R., Nobuhara, S., Nishino, K.: Polarimetric normal stereo.
  In: Proceedings of the IEEE/CVF Conference on Computer Vision and Pattern
  Recognition. pp. 682--690 (2021)

\bibitem{huynh2010shape}
Huynh, C.P., Robles-Kelly, A., Hancock, E.: Shape and refractive index recovery
  from single-view polarisation images. In: 2010 IEEE Computer Society
  Conference on Computer Vision and Pattern Recognition. pp. 1229--1236. IEEE
  (2010)

\bibitem{kadambi2015polarized}
Kadambi, A., Taamazyan, V., Shi, B., Raskar, R.: Polarized 3d: High-quality
  depth sensing with polarization cues. In: Proceedings of the IEEE
  International Conference on Computer Vision. pp. 3370--3378 (2015)

\bibitem{korger2013polarizer}
Korger, J., Kolb, T., Banzer, P., Aiello, A., Wittmann, C., Marquardt, C.,
  Leuchs, G.: The polarization properties of a tilted polarizer. Optics express
   \textbf{21}(22),  27032--27042 (2013)

\bibitem{Lei_2022_CVPR}
Lei, C., Qi, C., Xie, J., Fan, N., Koltun, V., Chen, Q.: Shape from
  polarization for complex scenes in the wild. In: Proceedings of the IEEE/CVF
  Conference on Computer Vision and Pattern Recognition (CVPR). pp.
  12632--12641 (June 2022)

\bibitem{miyazaki2012pol}
Miyazaki, D., Shigetomi, T., Baba, M., Furukawa, R., Hiura, S., Asada, N.:
  Polarization-based surface normal estimation of black specular objects from
  multiple viewpoints. In: 2012 Second International Conference on 3D Imaging,
  Modeling, Processing, Visualization \& Transmission. pp. 104--111. IEEE
  (2012)

\bibitem{miyazaki2012polarization}
Miyazaki, D., Shigetomi, T., Baba, M., Furukawa, R., Hiura, S., Asada, N.:
  Polarization-based surface normal estimation of black specular objects from
  multiple viewpoints. In: 2012 Second International Conference on 3D Imaging,
  Modeling, Processing, Visualization \& Transmission. pp. 104--111. IEEE
  (2012)

\bibitem{miyazaki2003polarization}
Miyazaki, D., Tan, R.T., Hara, K., Ikeuchi, K.: Polarization-based inverse
  rendering from a single view. In: Computer Vision, IEEE International
  Conference on. vol.~3, pp. 982--982. IEEE Computer Society (2003)

\bibitem{ngo2015shape}
Ngo~Thanh, T., Nagahara, H., Taniguchi, R.i.: Shape and light directions from
  shading and polarization. In: Proceedings of the IEEE conference on computer
  vision and pattern recognition. pp. 2310--2318 (2015)

\bibitem{rahmann2000polarization}
Rahmann, S.: Polarization images: a geometric interpretation for shape
  analysis. In: Proceedings 15th International Conference on Pattern
  Recognition. ICPR-2000. vol.~3, pp. 538--542. IEEE (2000)

\bibitem{rahmann2001reconstruction}
Rahmann, S., Canterakis, N.: Reconstruction of specular surfaces using
  polarization imaging. In: Proceedings of the 2001 IEEE Computer Society
  Conference on Computer Vision and Pattern Recognition. CVPR 2001. vol.~1,
  pp.~I--I. IEEE (2001)

\bibitem{shakeri2021icra}
Shakeri, M., Loo, S.Y., Zhang, H., Hu, K.: Polarimetric monocular dense mapping
  using relative deep depth prior. IEEE Robotics and Automation Letters
  \textbf{6}(3),  4512--4519 (2021)

\bibitem{smith2016linear}
Smith, W.A., Ramamoorthi, R., Tozza, S.: Linear depth estimation from an
  uncalibrated, monocular polarisation image. In: European Conference on
  Computer Vision. pp. 109--125. Springer (2016)

\bibitem{polsensor}
Sony: Polarization image sensor with four-directional on-chip polarizer and
  global shutter function,
  \url{https://www.sony-semicon.co.jp/e/products/IS/industry/product/polarization.html},
  accessed 2022-02-14

\bibitem{ting2021deep}
Ting, J., Wu, X., Hu, K., Zhang, H.: Deep snapshot hdr reconstruction based on
  the polarization camera. In: 2021 IEEE International Conference on Image
  Processing (ICIP). pp. 1769--1773. IEEE (2021)

\bibitem{tozza2017linear}
Tozza, S., Smith, W.A., Zhu, D., Ramamoorthi, R., Hancock, E.R.: Linear
  differential constraints for photo-polarimetric height estimation. In:
  Proceedings of the IEEE international conference on computer vision. pp.
  2279--2287 (2017)

\bibitem{tozza2021uncalibrated}
Tozza, S., Zhu, D., Smith, W., Ramamoorthi, R., Hancock, E.: Uncalibrated, two
  source photo-polarimetric stereo. IEEE Transactions on Pattern Analysis and
  Machine Intelligence  (2021)

\bibitem{wolff1990surface}
Wolff, L.B.: Surface orientation from two camera stereo with polarizers. In:
  Optics, Illumination, and Image Sensing for Machine Vision IV. vol.~1194, pp.
  287--297. SPIE (1990)

\bibitem{wu2020hdr}
Wu, X., Zhang, H., Hu, X., Shakeri, M., Fan, C., Ting, J.: Hdr reconstruction
  based on the polarization camera. IEEE Robotics and Automation Letters
  \textbf{5}(4),  5113--5119 (2020)

\bibitem{ACMM}
Xu, Q., Tao, W.: Multi-scale geometric consistency guided multi-view stereo.
  Computer Vision and Pattern Recognition (CVPR)  (2019)

\bibitem{yang2018}
Yang, L., Tan, F., Li, A., Cui, Z., Furukawa, Y., Tan, P.: Polarimetric dense
  monocular slam. In: Proceedings of the IEEE conference on computer vision and
  pattern recognition. pp. 3857--3866 (2018)

\bibitem{yu2017shape}
Yu, Y., Zhu, D., Smith, W.A.: Shape-from-polarisation: a nonlinear least
  squares approach. In: Proceedings of the IEEE International Conference on
  Computer Vision Workshops. pp. 2969--2976 (2017)

\bibitem{zhao2020polarimetric}
Zhao, J., Monno, Y., Okutomi, M.: Polarimetric multi-view inverse rendering.
  In: European Conference on Computer Vision. pp. 85--102. Springer (2020)

\bibitem{zhou2013multi}
Zhou, Z., Wu, Z., Tan, P.: Multi-view photometric stereo with spatially varying
  isotropic materials. In: Proceedings of the IEEE Conference on Computer
  Vision and Pattern Recognition. pp. 1482--1489 (2013)

\end{thebibliography}
\end{document}